\documentclass{article}

\usepackage{microtype}
\usepackage{graphicx}
\usepackage{subcaption}
\usepackage{booktabs} %

\usepackage[utf8]{inputenc} %
\usepackage[T1]{fontenc}    %
\usepackage{hyperref}       %
\usepackage{url}            %
\usepackage{booktabs}       %
\usepackage{amsfonts}       %
\usepackage{nicefrac}       %
\usepackage{microtype}      %
\usepackage{xcolor}         %

\usepackage{graphicx}
\usepackage{float}

\usepackage{hyperref}

\usepackage[preprint]{neurips_2024}

\renewcommand{\cite}[1]{\citep{#1}}

\usepackage{amsmath}
\usepackage{amssymb}
\usepackage{mathtools}
\usepackage{amsthm}

\usepackage[capitalize,noabbrev]{cleveref}

\usepackage{siunitx}

\theoremstyle{plain}

\theoremstyle{definition}

\theoremstyle{remark}

\DeclareMathOperator{\corr}{corr}

\usepackage[textsize=tiny]{todonotes}

\title{BrainWavLM: Fine-tuning Speech Representations with Brain Responses to Language}

\author{%
	Nishitha Vattikonda* \\
	Department of Computer Science\\
	The University of Texas at Austin\\
	\texttt{nishitha.vattikonda@utexas.edu} \\
	\And
	Aditya R. Vaidya* \\
	Department of Computer Science\\
	The University of Texas at Austin\\
	\texttt{avaidya@utexas.edu} \\
	\AND
	Richard J. Antonello \\
	Zuckerman Institute \\
	Columbia University \\
	\And
	Alexander G. Huth \\
	Departments of Computer Science and Neuroscience\\
	The University of Texas at Austin\\
	\texttt{huth@cs.utexas.edu} \\
}

\begin{document}

\maketitle

\begin{abstract}
Speech encoding models use auditory representations to predict how the human brain responds to spoken language stimuli. Most performant encoding models linearly map the hidden states of artificial neural networks to brain data, but this linear restriction may limit their effectiveness. In this work, we use low-rank adaptation (LoRA) to fine-tune a WavLM-based encoding model end-to-end on a brain encoding objective, producing a model we name BrainWavLM.
We show that fine-tuning across all of cortex improves average encoding performance with greater stability than without LoRA. This improvement comes at the expense of low-level regions like auditory cortex (AC), but selectively fine-tuning on these areas improves performance in AC, while largely retaining gains made in the rest of cortex.
Fine-tuned models generalized across subjects, indicating that they learned robust brain-like representations of the speech stimuli.
Finally, by training linear probes, we showed that the brain data strengthened semantic representations in the speech model without any explicit annotations.
Our results demonstrate that brain fine-tuning produces best-in-class speech encoding models, and that non-linear methods have the potential to bridge the gap between artificial and biological representations of semantics.
\end{abstract}

\section{Introduction}
\label{submission}

Many neuroscience experiments record brain responses while participants listen to natural language stimuli, such as narrative stories~\cite{wehbeAligningContextbasedStatistical2014,huthNaturalSpeechReveals2016,nastaseNarrativesFMRIDataset2021}. These data are used to fit encoding models that predict the response at each location in the brain as a function of the stimulus~\cite{naselarisEncodingDecodingFMRI2011}. Estimated encoding models can then be used to study which types of information are represented in each brain area and many other questions~\cite{goldsteinBrainEmbeddingsShared2022,ivanovaLinearRegressionMapping2022,caucheteuxEvidencePredictiveCoding2023,tangSemanticReconstructionContinuous2023,weissbartStructureStatisticsLanguage2024}.

Most encoding models take the form of \textit{linearized models}, where a pre-trained non-linear function is first used to extract features from the stimuli, and then a linear mapping is learned from those features to the brain data~\cite{wuCompleteFunctionalCharacterization2006}. Linearized models are very efficient, making them a good fit for limited neuroscience datasets. But as these datasets grow~\cite{lebelNaturalLanguageFMRI2023}, there is an increasing possibility that fully non-linear encoding models like deep artificial neural networks (ANNs) could perform better than linearized models.

Here we explore a hybrid approach in which a well-performing linearizing transform, WavLM~\cite{chenWavLMLargeScaleSelfSupervised2021}, is fine-tuned to predict brain data. To make the best use of fMRI datasets that are large for neuroscience but small compared to typical neural network pre-training datasets, we also lean on low-rank adaptation (LoRA; \citealp{huLoRALowRankAdaptation2021}) methods for doing efficient fine-tuning. We found that fine-tuned models yield significantly better prediction performance on held-out test data than linearized models, demonstrating the power of these techniques and datasets.

We then used this approach to test scientific questions: Are the features learned by nonlinear encoding models generalizable across different brain areas, hemispheres, and people? How do the fine-tuned representations differ from the base model? We found surprisingly robust generalization in each case, and showed that semantic representations seem enhanced at the expense of acoustic representations. These results demonstrate the power of non-linear encoding models in neuroscience. They also raise the possibility that neural network models for speech or language understanding could be trained directly on brain data, providing a more powerful supervisory signal than existing pre-training approaches.

\section{Data and Methods}

\subsection{fMRI Data}

From the public dataset released by \citet{lebelNaturalLanguageFMRI2023} and \citet{tangSemanticReconstructionContinuous2023}, we used pre-processed fMRI data from 3 participants who were scanned while listening to 94--103 natural stories (\qtyrange[range-phrase=--]{17.8}{19.7}{\hour} hours per participant).
These stories capture a range of semantics that could be useful to capture in an ANN.

During scanning, whole-brain volumes were captured every 2 seconds (TR = \SI{2}{\second}) at \SI{2.6}{\mm} isotropic resolution. Acquisition and pre-processing procedures are described in detail in \citet{lebelNaturalLanguageFMRI2023}. To improve signal quality, responses for three stories were averaged across multiple presentations. During fine-tuning, one story (``wheretheressmoke'') was used as a test set and two (``fromboyhoodtofatherhood'' and ``onapproachtopluto'') were used as a validation set. All remaining stories were presented once and were used for training.

\subsection{Speech Encoding Models}
\label{sec:methods-encmodels}

Our work focuses on encoding models that operate on the audio of the stimulus to predict responses~\cite{vaidyaSelfSupervisedModelsAudio2022,milletRealisticModelSpeech2022,tuckuteManyNotAll2023,ootaSpeechLanguageModels2024}. We first fit linearized encoding models to obtain a baseline for encoding performance. These models take the form $f(S) = g(S) \beta$, where $S$ is a stimulus (in our case, spoken language), $g$ is a frozen $P$-dimensional non-linear transform of the stimulus, and $\beta \in \mathbb{R}^{P \times |V|}$ is a linear transform between the output space of $g$ and the response for voxels $V$.
They aim to estimate true responses $R \in \mathbb{R}^{T \times |V|}$, where $T$ is the number of fMRI volumes.
We used pre-trained audio-based ANNs for $g$, and independently estimate $f_v$ for each voxel $v \in V$.

We used previously published procedures to produce aligned speech features for every fMRI volume~\cite{vaidyaSelfSupervisedModelsAudio2022,antonelloScalingLawsLanguage2023}. We moved a sliding window with size \SI{2.0}{\second} and stride \SI{0.1}{\second} over the audio stimulus $S$ and fed the result into an audio model. We saved the hidden state at the final token of each layer. We interpolated the result to the capture time of the fMRI volumes with a Lanczos filter.

Given aligned speech features $g(S) \in \mathbb{R}^{T \times P}$ and fMRI responses $R \in \mathbb{R}^{T \times |V|}$, we used ridge regression to estimate encoding weights $\beta$ on delayed speech features of the training set. Features were delayed by 2, 4, 6, and 8 seconds to account for the hemodynamic response function~\cite{nishimotoReconstructingVisualExperiences2011,huthNaturalSpeechReveals2016}. Encoding performance $\rho_v$ was calculated for each voxel $v$ as:
\begin{equation} %
\label{eq:encperf}
    \rho_v = \corr_t \left( R_{:,v}, \hat{R}_{:,v} \right)
\end{equation}
where $\corr_t$ denotes the correlation over time, and $R_{:,v}$ and $\hat{R}_{:,v}$ are the true and predicted timecourses, respectively, for voxel $v$.

\subsection{Fine-tuning Deep Neural Networks with Brain Data}
\label{sec:methods-lora}

We adapted the standard encoding model fitting procedure to be trainable end-to-end. Our procedure fine-tunes a BrainWavLM network, parametrized by weights $\theta_g$, to predict fMRI responses given the audio stimuli that elicited them. Brain responses were predicted by a linear projection, parametrized by $\theta_p$, from the hidden state of one layer of the network. We applied the same feature extraction and downsampling procedure as we do for the non-finetuned models in \cref{sec:methods-encmodels}. We then optimized $(\theta_g, \theta_p)$ to minimize the loss function:
\begin{equation} %
\label{eq:loss}
    \mathcal{L}(\theta_g, \theta_p) =  -\frac{1}{T} \sum_{t=1}^{T} \corr_v \left( R_{t,:}, \hat{R}_{t,:} \right)
\end{equation}
where $\corr_v$ denotes the \emph{spatial} correlation (i.e., over voxels), and $R_{t,:}$ and $\hat{R}_{t,:}$ are the true and predicted fMRI volumes, respectively, at time $t$. Importantly, this differs from the measure for encoding performance~\eqref{eq:encperf}, which uses the temporal correlation, because we found the backpropagation to be faster.

\subsubsection{Training Details}
\label{sec:methods-lora-training}

The architecture of BrainWavLM, the backbone of our encoding models, follows that of WavLM Base~\cite{chenWavLMLargeScaleSelfSupervised2021}, and we initialize its weights with the public WavLM Base+ checkpoint.
We apply the linear projection to the hidden states of the 9th layer, since it has been shown to be the layer with the highest encoding performance among similarly sized models~\cite{antonelloScalingLawsLanguage2023}.
Rather than optimizing every parameter of the BrainWavLM model, we used LoRA to learn rank-$4$ updates on the $W^Q$, $W^K$, and $W^V$ matrices of each Transformer layer. We also constrain the linear projection to be low-rank via a rank-$100$ bottleneck. (We found that the first 100 principal components of the pre-trained model's encoding weights capture more than 70\% of the variance across voxels.)

We used the Adam optimizer~\cite{kingmaAdamMethodStochastic2017} with learning rate of \num{1e-4}. We only optimized the LoRA and final linear bottleneck --- the rest of the BrainWavLM model, including WavLM's convolutional feature encoder, were frozen.
In total, we have 68M frozen parameters through the first 9 layers of BrainWavLM, and we have 166K learnable LoRA parameters.
The number of learnable parameters in the final linear bottleneck varies by participant, between 8.5M and 9.9M.

We fine-tuned the models for 20 epochs with a batch size of 50 TRs.
Fine-tuning a single model took about $\SI{2}{\hour}$ on one NVIDIA A100 40GB GPU.

\subsubsection{Evaluating Fine-tuned Models}
\label{sec:methods-lora-eval}

If we wish to stop fine-tuning when the validation encoding performance converges, we cannot simply halt training when validation \emph{loss} stops decreasing, since the training objective~\eqref{eq:loss} and encoding performance~\eqref{eq:encperf} are defined differently. Instead, at every epoch, we discard the final linear projection and re-fit a linearized encoding model using the procedure from \cref{sec:methods-encmodels}. We then choose the epoch with the best validation encoding performance. All figures show the best-performing epoch's encoding performance on the test set, unless otherwise stated.

\subsubsection{Baseline Fine-tuning Models}

We train two additional models as baselines for our LoRA fine-tuning method. First, to test whether the brain data contains semantic information that would otherwise be learnable from word annotations, we fine-tuned one model to predict features from LLaMA 33B~\cite{touvronLLaMAOpenEfficient2023}, a large language model that has been previously shown to capture a breadth of responses in high-level brain areas~\cite{antonelloScalingLawsLanguage2023}. We replaced the fMRI responses with hidden states from layer 18, which had the highest encoding performance, by feeding the stimulus transcripts into the model and downsampling the outputs to match the fMRI acquisition rate. After fine-tuning was complete, we fit and tested cortical encoding models normally using the new LLaMA-tuned features.
If fine-tuning on fMRI data outperforms this baseline, it indicates that the brain responses contain information that is absent even in existing rich representations.

We also tested the importance of LoRA in our fine-tuning procedure by fine-tuning another model without LoRA. Following~\citet{moussaImprovingSemanticUnderstanding2024}, we updated all weights of the BrainWavLM model during optimization and used a learning rate schedule.

\subsection{Linguistic Representation Probes}
\label{sec:methods-probes}

To quantitatively describe how fine-tuning alters model representations, we probed the representations with known linguistic features.
We used linear probes to match the expressivity of the probe with that of a linear encoding model~\cite{yangSUPERBSpeechProcessing2021,pasadLayerwiseAnalysisSelfsupervised2021}.

We first extracted layer-wise features of an audio stimulus using the sliding window approach from \cref{sec:methods-encmodels}, without downsampling. To probe acoustic content, we filtered the stimulus waveform with a filterbank~\cite{poveyKaldiSpeechRecognition2011,yangTorchAudioBuildingBlocks2022}, and we used ridge regression to linearly predict the resulting features.

We used GloVe embeddings~\cite{penningtonGloveGlobalVectors2014} to probe for word-level semantic content following \citet{vaidyaSelfSupervisedModelsAudio2022}. With the time-aligned transcripts from the original dataset, we aligned the features extracted from the BrainWavLM model to the words being spoken.
We then used ridge regression to predict the 300-D GloVe embedding of the word being spoken. We measured the performance of both probes with variance explained ($R^2$).

All probes were trained on 26 stories (\SI{5.4}{\hour}) and tested on the 3 stories that are held out during brain fine-tuning.
We trained separate probes on each layer of the fine-tuned model, as well as the pre-trained model to measure baseline performance.

\section{Results}

\subsection{Fine-tuning on All of Cortex}
\label{sec:results-base}

\begin{figure*}[ht!]
    \centering
    \includegraphics[width=1\linewidth]{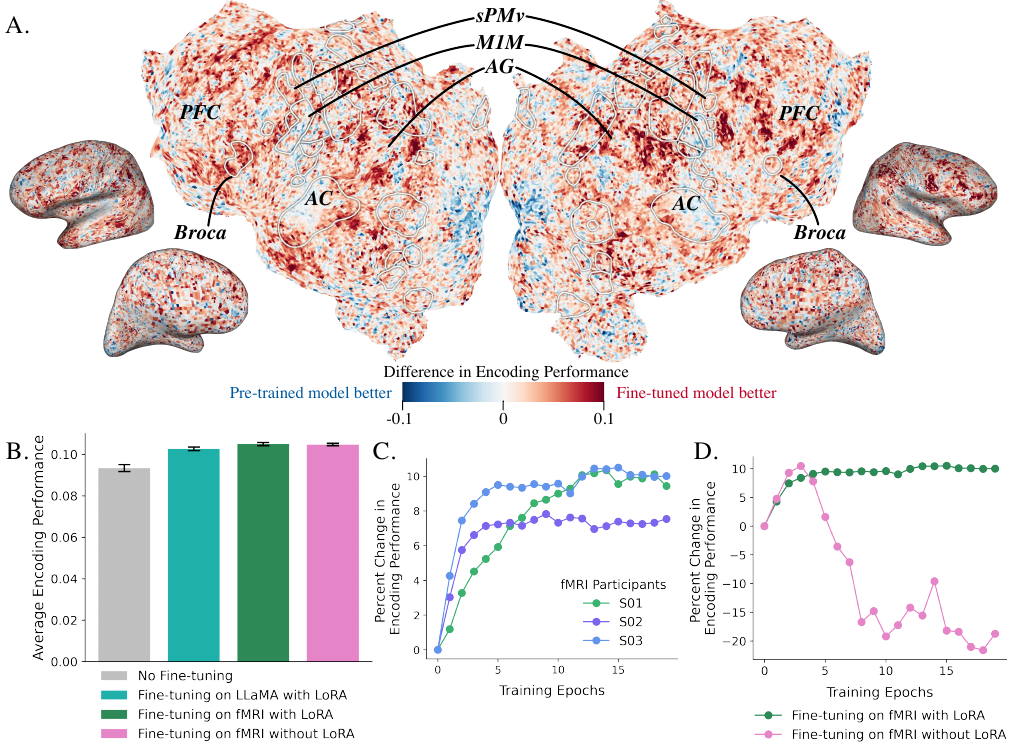}
    \caption{\textbf{Encoding performance of BrainWavLM models fine-tuned on fMRI responses.} (A) Cortical map of the change in encoding performance from the pre-trained WavLM model to the highest-performing BrainWavLM model (measured by performance on the validation set) on the test set. Corresponds to model fine-tuned with LoRA in (B). Results shown are for subject S03. (B) Encoding performance on the test set for the pre-trained model, model fine-tuned on LLaMA features with LoRA, model fine-tuned on fMRI data with LoRA, and model fine-tuned on fMRI data without LoRA, averaged across voxels and then subjects. Error bars show the standard error of the mean (SEM) for the per-subject performance, corrected for the overall performance by subtracting each subject's mean performance across the models from each model's performance.
    (C) Percent change in validation encoding performance through model fine-tuning with LoRA. The biggest improvements are found in the first 10 epochs, after which the performance stabilizes. (D) Change in validation encoding performance for the models fine-tuned with and without LoRA on subject S03. Performance for the model fine-tuned with LoRA was more stable during training. Validation performance for additional subjects is shown in Fig.~\ref{fig:app-subject-valperf} in \cref{app:subject-valperf}.}
    \label{fig:lora-encperf}
\end{figure*}

We fine-tuned separate models on responses from each fMRI participant.
This caused average encoding performance across voxels to improve by 12.5\% above the pre-trained WavLM Base+ model (Fig.~\ref{fig:lora-encperf}B).
Fine-tuning on brain data outperformed fine-tuning on LLaMA features (11.1\% improvement over pre-trained), indicating that the brain data may have more relevant semantic content.

While average encoding performance across cortex increased, we saw a notable drop in early auditory cortex (AC) and other low-level areas (Fig.~\ref{fig:lora-encperf}A), similar to what was found by \citet{moussaImprovingSemanticUnderstanding2024}. Different fMRI participants required different amounts of training to reach maximum validation performance, but performance plateaued and does not decrease after this point (Fig.~\ref{fig:lora-encperf}C).

Our LoRA-based method had comparable performance to the full-model fine-tuning of \citet{moussaImprovingSemanticUnderstanding2024} (Fig.~\ref{fig:lora-encperf}B), suggesting that a low-rank update may be all that is needed. LoRA-based fine-tuning was also more stable (Fig.~\ref{fig:lora-encperf}D), but took more epochs to reach the same validation performance.

\subsection{Fine-tuning on Subsets of Cortex}
\label{sec:results-rois}

\begin{figure*}[ht!]
    \centering
    \includegraphics[width=1\linewidth]{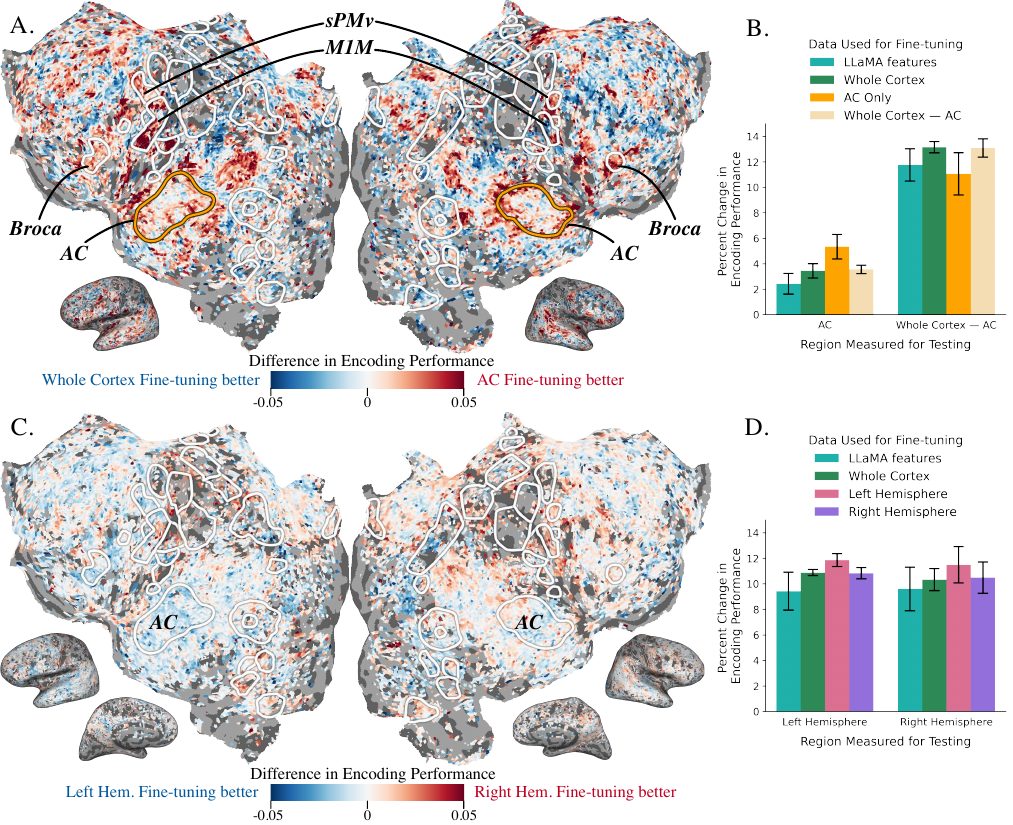}
    \caption{\textbf{Models adapt to representations in subsets of cortex.} (A) Encoding performance was computed for models fine-tuned on the whole cortex or fine-tuned just on auditory cortex (AC). Cortical maps show the difference in encoding performance on one subject. Only voxels with encoding performance above $0.15$ for the pre-trained model are shown. The model fine-tuned on AC has higher performance in language-selective areas in the temporal and frontal lobes~\cite{fedorenkoFunctionalSpecificityHighlevel2011,lipkinProbabilisticAtlasLanguage2022}. (B) Average percent improvement in encoding performance from the pre-trained WavLM Base+ model to the fine-tuned BrainWavLM models, either computed within AC or across the rest of cortex. One model was fine-tuned to predict features from LLaMA, and three models were fine-tuned on fMRI responses from either the whole cortex (81K--95K voxels), only auditory cortex (1.3K--2.7K voxels), or the whole cortex except auditory cortex (79K--93K voxels). Error bars show the SEM for the per-subject performance. For both brain areas, fMRI-tuned models were better at predicting fMRI responses than the LLaMA-tuned model or the pre-trained WavLM model. The best model for AC was fine-tuned on AC, and the best model for the rest of cortex was fine-tuned on the rest of cortex.
    (C) Separate models were fine-tuned using only left- or right-hemisphere voxels. Cortical maps show the difference in performance. Voxels are filtered with the same condition as in (A). Similar maps for other subjects are shown in Fig.~\ref{fig:app-hemis-flatmaps} in \cref{app:roi-flatmaps}.
    (D) Percent improvement in encoding performance from the pre-trained model to the left- and right-hemisphere fine-tuned models, with models from (B) for comparison. Left-hemisphere fine-tuned models were marginally better across cortex than those fine-tuned on the right-hemisphere, though both were substantially better than the pre-trained model.}
    \label{fig:roi-encperf}
\end{figure*}

One puzzling aspect of these results is that model performance after fine-tuning increases across the board but slightly decreases in auditory cortex (AC), an area that we expect to have excellent signal quality \cite{antonelloScalingLawsLanguage2023}. We hypothesized that this could be due to competing demands between AC and higher-level semantic brain areas during fine-tuning. The much larger fraction of voxels tuned to semantic information \cite{antonelloScalingLawsLanguage2023} may simply dominate the fine-tuning loss, to the detriment of AC. To test this hypothesis, we fine-tuned two complementary models for each subject: one that only learns from responses for voxels within AC (as localized by~\citet{lebelNaturalLanguageFMRI2023}), and one that uses all voxels except those in AC. Following the procedure from \cref{sec:methods-lora-eval}, we then fit and evaluated whole-cortex encoding models using each fine-tuned model.

We compared these models against the original models fine-tuned on all of cortex, looking at performance within and outside AC (Fig.~\ref{fig:roi-encperf}A). The models fine-tuned on non-AC voxels performed comparably to the model fine-tuned on all of cortex, with no significant change within or outside of AC. Models fine-tuned only on AC, however, performed better within AC (5.5\% vs. 4\%) and worse outside AC (11\% vs. 13\%). Though performance outside AC was hurt slightly, it was still a significant improvement over the pre-trained model. In addition to AC, performance increased in other low-level areas like M1M (primary mouth motor area) and sPMv (ventral speech pre-motor area) (Fig.~\ref{fig:roi-encperf}B). These results suggest that the features learned by fine-tuning on a low-level area like AC generalize to higher-level brain areas as well.

With our fine-tuning procedure, we can ask other questions about the nature of representations in different parts of the brain. One interesting question is how representations of speech and language differ between the left and right hemispheres~\cite{hickokCorticalOrganizationSpeech2007,albouyDistinctSensitivitySpectrotemporal2020,ozernov-palchikPrecisionFMRIReveals2024}. Our approach makes it possible to test whether the linguistic features learned by fine-tuning on brain data are able to generalize from each hemisphere to the other. For each subject, we fine-tuned separate models on each hemisphere of the brain and fit whole-cortex encoding models with the resulting features. If the hemispheres had fundamentally different functions, we would expect, for example, that a model fine-tuned only on the left hemisphere would not improve in the right hemisphere.

\begin{figure*}[ht!]
    \centering
    \includegraphics[width=1\linewidth]{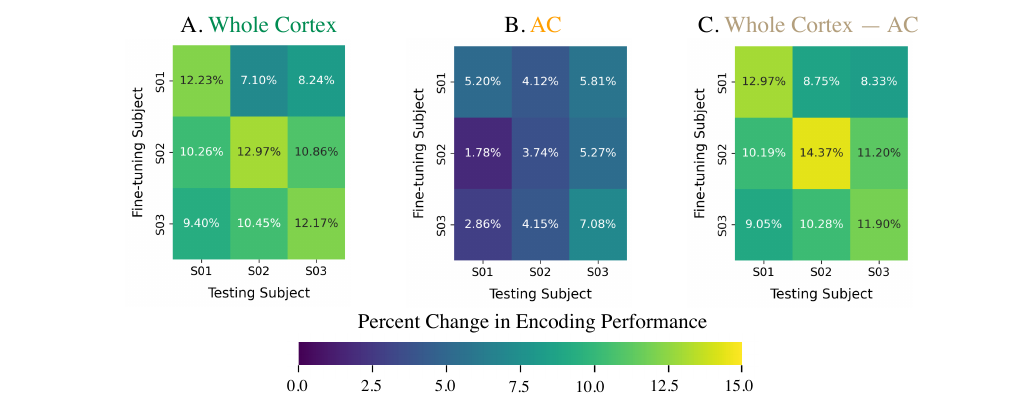}
    \caption{\textbf{Fine-tuned models transfer between fMRI subjects.} Percent improvement in encoding performance (averaged across voxels) of the models fine-tuned using one subject's fMRI responses compared to the pre-trained model. (A) Models were fine-tuned using fMRI responses from the whole cortex. (B) Models were fine-tuned using fMRI responses from auditory cortex (AC). Performance was only measured on the voxels within auditory cortex. (C) Models were fine-tuned using fMRI responses from the whole cortex except AC. Performance was only measured on the voxels outside auditory cortex. In all cases we see substantial generalization between subjects, with improvements of 7--14\% outside AC but only 1--6\% inside AC.}
    \label{fig:transfer-encperf}
\end{figure*}

Overall, we found that fine-tuning on either hemisphere improved model performance by a comparable amount, with a slightly larger improvement from fine-tuning on the left rather than right hemisphere (Fig.~\ref{fig:roi-encperf}C-D).
Models fine-tuned on the right hemisphere performed similarly to those fine-tuned on all of cortex. This may reflect small differences in the extent of language-selective cortex in each hemisphere~\cite{deheerHierarchicalCorticalOrganization2017}, but overall the high cross-hemisphere transfer performance suggests that the representations learned during fine-tuning are useful bilaterally.

\subsection{Transferring Across Subjects}
\label{sec:results-subjects}

Fine-tuned models are able to improve performance within subjects, so we next investigated whether fine-tuning on one subject would improve encoding performance in another subject. This would indicate that the learned features were generalizable across subjects.

We extracted features from the best model fine-tuned on each training fMRI subject in \cref{sec:results-base}, and used the features to fit encoding models that predict the responses for each other test subject. As in previous experiments, we compared the performance of these encoding models against that of an encoding model that used the pre-trained WavLM model. We measured each model's transfer performance by averaging its performance improvement on other test subjects.

Fine-tuned models from any training subject improved encoding performance on all test subjects by at least 7\% over the pre-trained model (Fig.~\ref{fig:transfer-encperf}A), indicating that all fine-tuned models learned generalizable features. Nonetheless, cross-subject transfer performance was worse than within-subject performance (i.e., encoding improvement was highest when fine-tuning and testing on the same subject). Training on subjects S02 and S03 yielded overall better transfer performance than with S01 (10.56\% and 9.93\%, vs. 7.67\%, averaged across testing subjects), despite within-subject performance being similar. Regardless of the subject the model was fine-tuned on, performance on other testing subjects only deviated by about 1\%.

With the same methodology, we additionally tested the models fine-tuned on AC and the rest of cortex from \cref{sec:results-rois}. Fine-tuning on either partition yielded improvements in encoding performance, even after transferring the models across subjects. Transfer performance for LoRA weights fine-tuned on all of cortex except AC followed the trends of the cortex at large (Fig.~\ref{fig:transfer-encperf}C).

\begin{figure*}[ht!]
    \centering
    \includegraphics[width=1\linewidth]{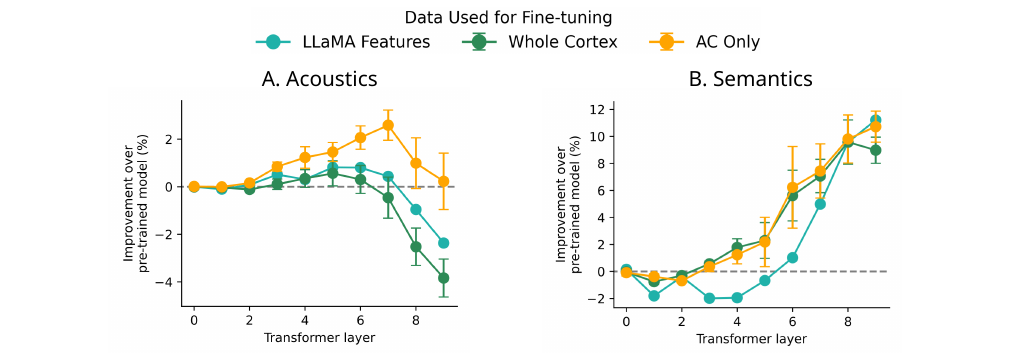}
    \caption{\textbf{Model representations change after fine-tuning.} Probe improvement over the pre-trained model, averaged across subjects. Transformer layer 0 is WavLM's convolutional waveform encoder. Error bars indicate SEM across subjects. (A) Acoustic probes linearly predict filterbank features of the stimulus waveform. Models fine-tuned on the whole cortex or on LLaMA became less acoustic, whereas the middle layers of AC-tuned models became more acoustic. (B) Semantic probes linearly predict GloVe embeddings of the time-aligned transcript. GloVe probes had the same performance for fMRI-tuned and LLaMA-tuned models, suggesting that the fMRI data is an effective source of semantics. Un-averaged probe performance can be seen in Fig.~\ref{fig:app-subject-probing} in \cref{app:subject-probing}.}
    \label{fig:probing}
\end{figure*}

Within AC, models had lower cross-subject transfer performance when testing on S01 than on S02 and S03 (Fig.~\ref{fig:transfer-encperf}B), indicating that the auditory functions fine-tuned to predict AC in S02 and S03 were more similar to each other than to S01. Transfer performance was positive in S01, however, so there is still some overlap. In contrast with our whole-cortex findings, to maximize a subject's encoding performance in AC, it may be optimal to fine-tune BrainWavLM on a different subject's responses. For example, a model fine-tuned with the AC of S02 had a greater improvement in S03 than in S02.

This shows that fMRI fine-tuned models largely transfer across subjects, and the magnitude of the transfer performance closely mirrors the within-subject performance. These models must be learning features that provide general utility for predicting fMRI responses.

\subsection{Probing Fine-tuned Representations}

What kinds of linguistic information are these ANNs learning from the brain data, and from what brain areas?

We turned to linear probes to answer these questions. We quantified the acoustic and semantic content in the model's representations by predicting the filterbank features and GloVe word embeddings of the original stimulus.
These probes were trained and evaluated separately for each layer of every fine-tuned and pre-trained WavLM model (see \cref{sec:methods-probes}). We only examined probe performance through the first 9 Transformer layers, since we used the output of layer 9 to predict brain data.

Fine-tuning on all of cortex reduced acoustic probing performance in later layers relative to the pre-trained model (Fig.~\ref{fig:probing}A), similar to the result of fine-tuning on features from LLaMA.
We believe these trends align because whole-cortex responses and LLaMA features are both primarily capturing semantics, but the effect is nonetheless stronger for the models fine-tuned on brain data.
As expected, models fine-tuned only on AC had enhanced acoustic probe performance compared to the other models.

While acoustic content remained similar or weakened by layer 9, fine-tuning greatly strengthened semantic representations compared to the pre-trained model, with larger shifts in deeper layers (Fig.~\ref{fig:probing}B).
Though models fine-tuned on brain data were comparable to LLaMA-tuned models by the output layer, the former show a more uniform increase in semantic probe performance across layers.

Despite having no annotations on word boundaries or meaning, fine-tuning on brain data strengthens semantic representations in WavLM at a level comparable to explicit semantic supervision through a language model.
Depending on which part of cortex is used for fine-tuning, this change can come at the cost of acoustic representations.

\section{Conclusion}

We developed a procedure to fine-tune WavLM-based speech encoding models on brain data using LoRA. These models outperformed previous linear models~\cite{antonelloScalingLawsLanguage2023}, though they underperformed in low-level areas. We believed this was because these areas are only a small portion of cortex and were overpowered by the more numerous semantic voxels. We resolved this discrepancy by fine-tuning only on AC, which created stronger low-level models that still generalized to the rest of cortex. With a similar procedure, we also tested for language lateralization in our data by fine-tuning on one hemisphere at a time. Within- and cross-hemisphere encoding performance was similar, suggesting that the functional selectivity of the hemispheres may be similar as well.

Since these adaptations transferred well across multiple brain regions of a single subject, we further stressed their generality by transferring them across subjects. These models beat the pre-trained baseline, even though they were fine-tuned on different subjects than they were tested on. This trend also held within AC, though this improvement was smaller than in the rest of cortex. We finally turned to linguistic probes to concretely describe how the models' representations changed after fine-tuning. The brain data caused the final layers of the speech model to become more semantic and less acoustic.

Our primary results are corroborated by \citet{moussaImprovingSemanticUnderstanding2024}, whose work on the same dataset shows that fine-tuning on fMRI data improves encoding performance outside of auditory cortex, and that the procedure biases model representations to be more semantic. Our paper extends the results of \citet{moussaImprovingSemanticUnderstanding2024} in two ways. First, we show that a loss in low-level encoding performance is not intrinsic to the brain-tuning procedure; it is instead a consequence of these voxels being outnumbered by semantic voxels. Auditory cortex encoding performance improves if that region is targeted during fine-tuning. Second, we show that models learn representations that improve performance in brain regions and subjects that they were not fine-tuned on. This result is complementary to the results on downstream tasks shown in their paper.

Overall, we demonstrate that fine-tuning speech encoding models on brain data improves encoding performance, both within and across subjects, and re-orients outputs to be more semantic. Further, this was done without any manual labels or annotations, since the WavLM model was trained via self-supervision and the fMRI task is passive listening. Our findings demonstrate the power of brain data for building robust semantic representations in ANNs. We will publicly release code and LoRA weights for fine-tuned BrainWavLM models after publication.

\setcitestyle{authoryear,open={((},close={))}} %
\bibliography{bib/arxiv}
\bibliographystyle{abbrvnat}

\newpage
\appendix
\onecolumn
\section{Validation encoding performance during fine-tuning}
\label{app:subject-valperf}

\begin{figure}[H]
    \centering
    
    \begin{subfigure}{0.4\linewidth}
    \includegraphics[width=\linewidth]{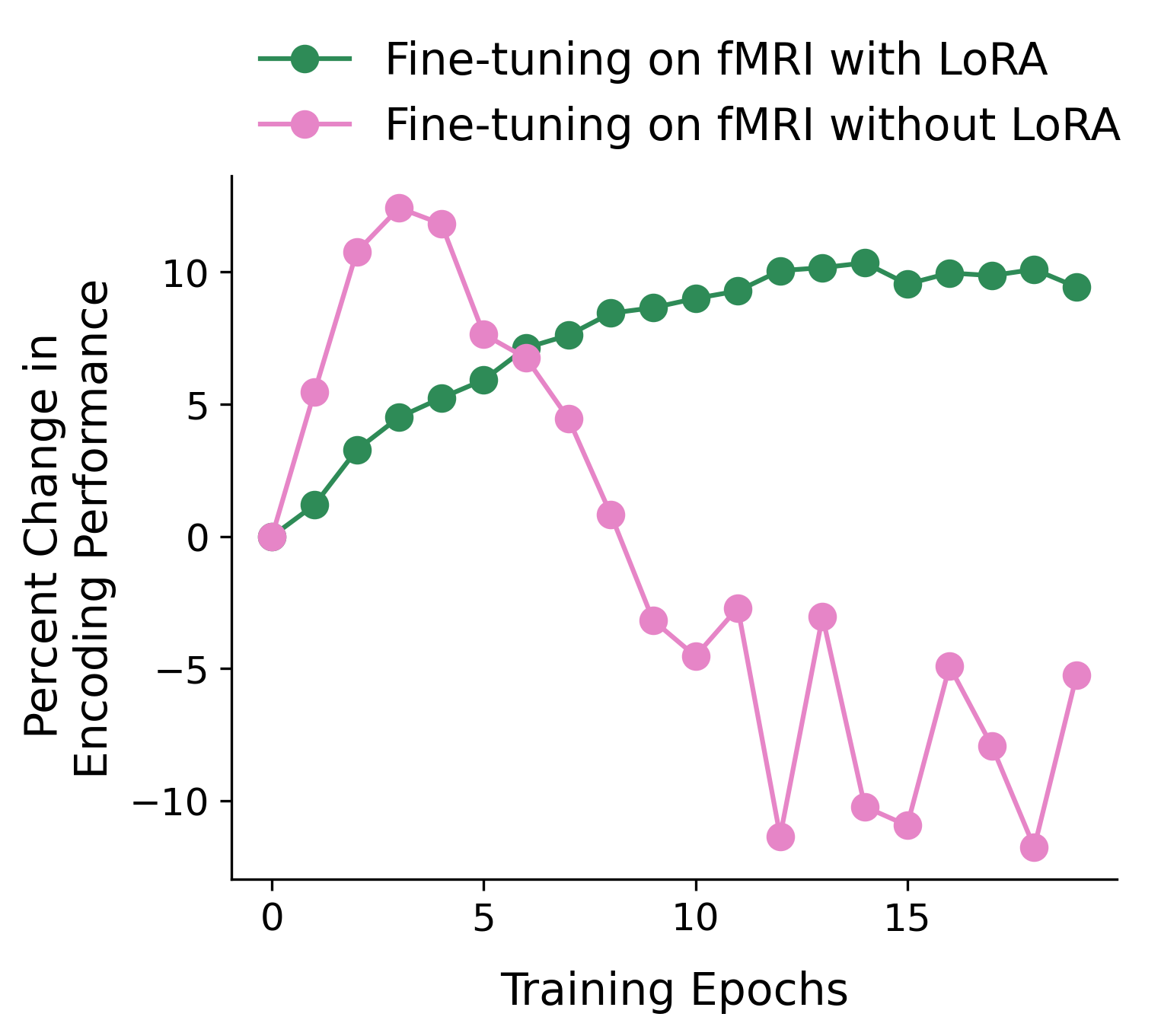} 
    \caption{S01}
    \end{subfigure}
    \hfill
    \begin{subfigure}{0.4\linewidth}
    \includegraphics[width=\linewidth]{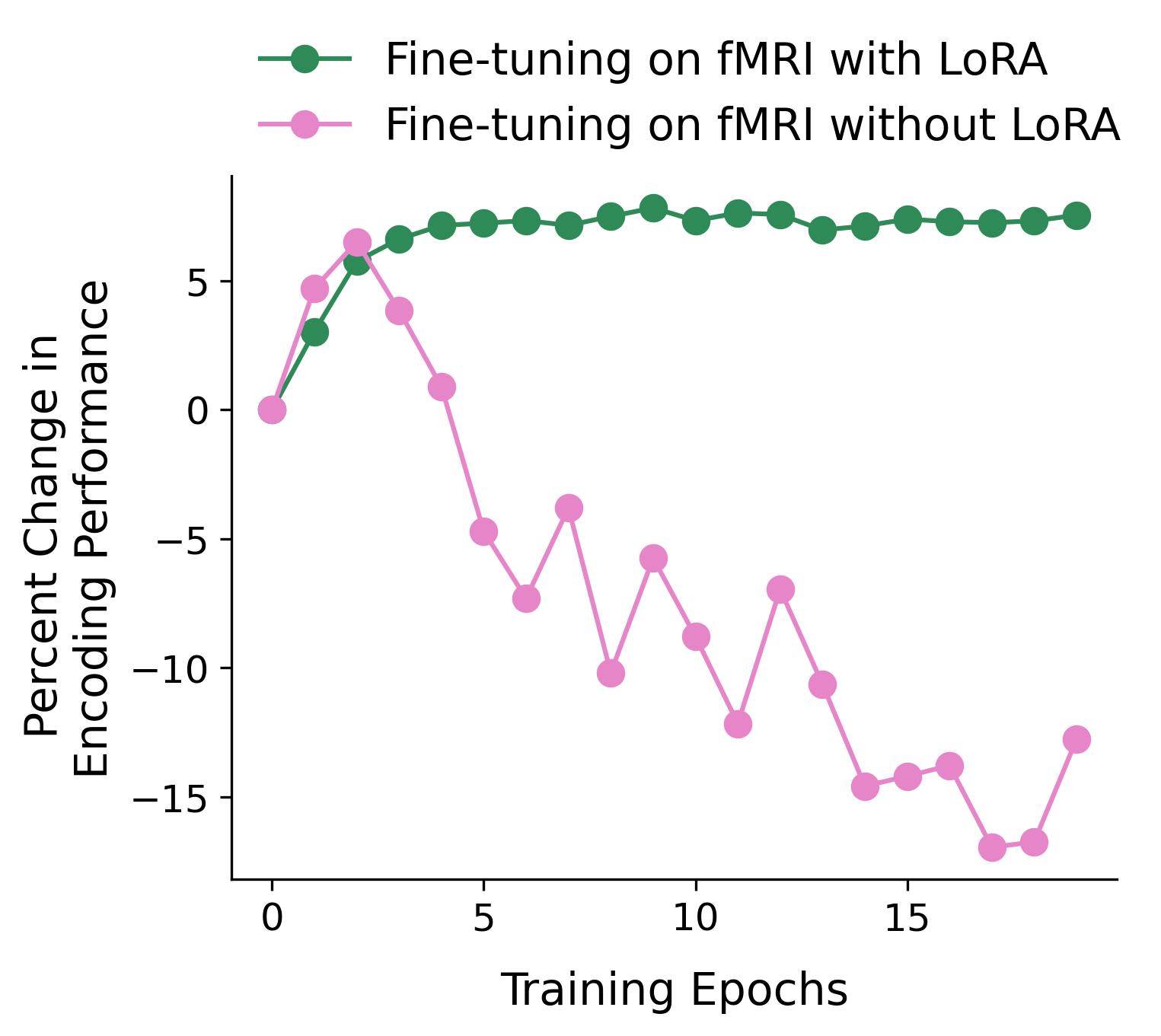} 
    \caption{S02}
    \end{subfigure}
    
    \caption{Percent Improvement in Encoding Performance for Models Fine-tuned with and without LoRA for subjects S01 and S02. Subject S03 is shown in Fig.~\ref{fig:lora-encperf}D.}
    \label{fig:app-subject-valperf}
\end{figure}

In Fig.~\ref{fig:app-subject-valperf}, we show the validation encoding performance for the two subjects not shown in the main text.

\newpage
\section{Fine-tuning on subsets of cortex}
\label{app:roi-flatmaps}

This section contains the cortical maps from Fig.~\ref{fig:roi-encperf} for subjects S01 and S02.

\begin{figure}[H]
    \centering
    \begin{subfigure}{0.8\linewidth}    \includegraphics[width=\linewidth]{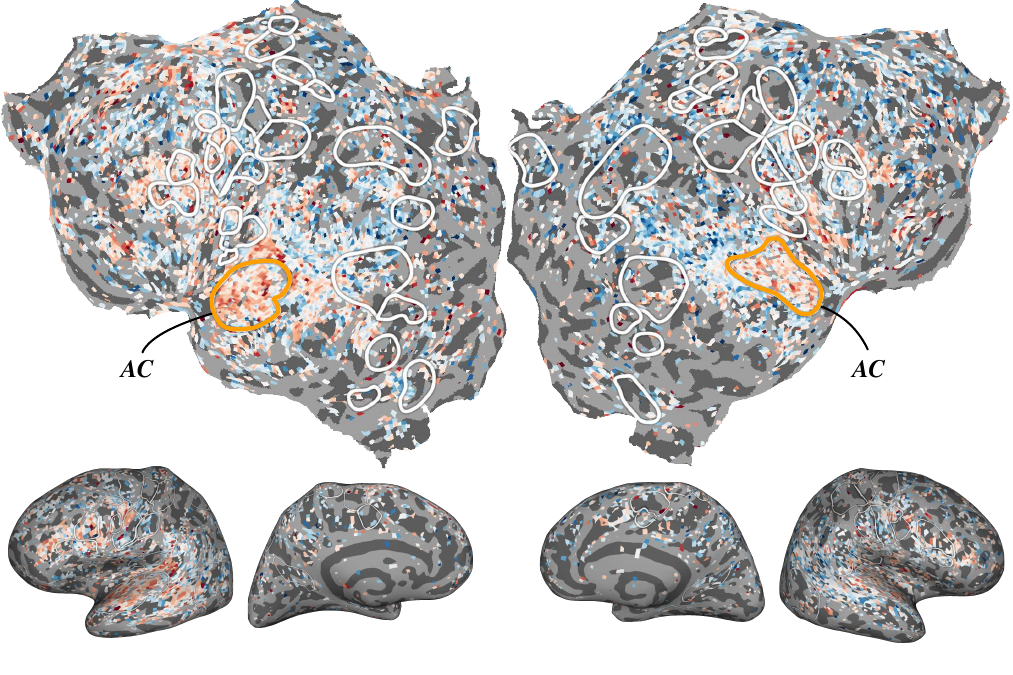} 
    \caption{S01}
    \end{subfigure}

    \begin{subfigure}{0.8\linewidth}    \includegraphics[width=\linewidth]{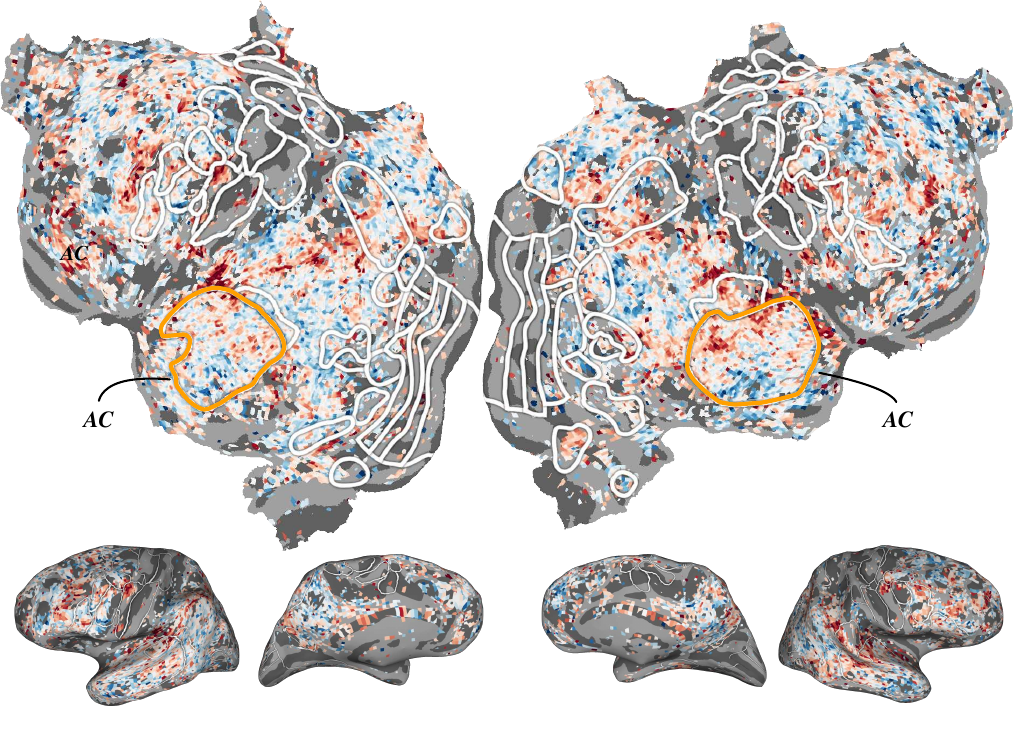} 
    \caption{S02}
    \end{subfigure}

    \caption{AC Fine-tuning Improvement over Whole Cortex Fine-tuning for subjects S01 and S02. Subject S03 is shown in Fig.~\ref{fig:roi-encperf}A.}
    \label{fig:app-ac-flatmaps}
\end{figure}

\begin{figure}[H]
    \centering
    \begin{subfigure}{0.8\linewidth}
    \includegraphics[width=\linewidth]{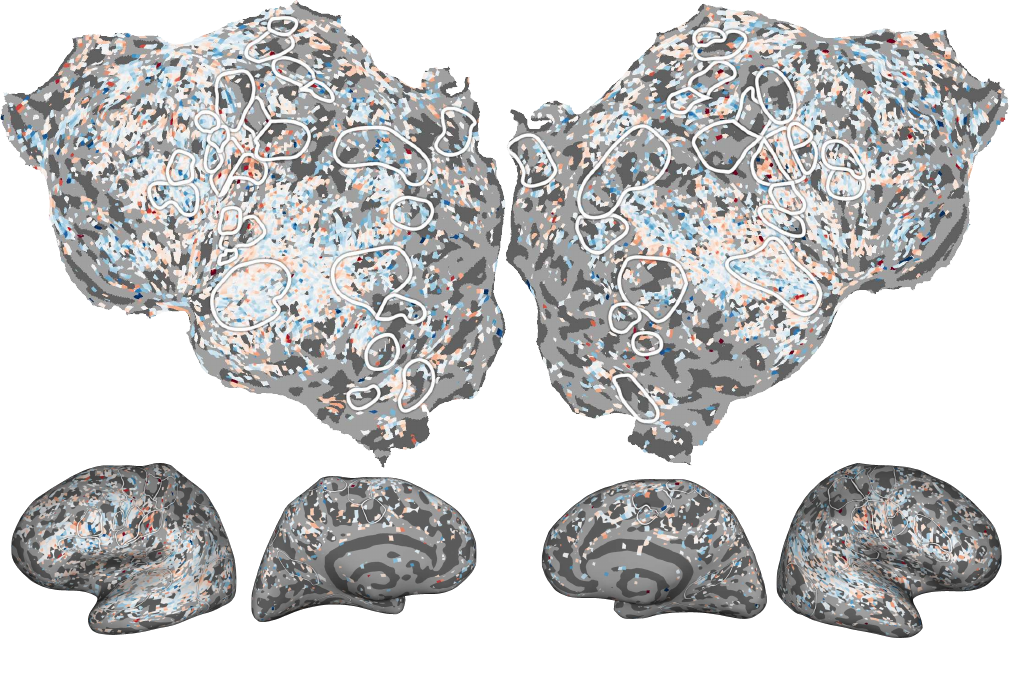} 
    \caption{S01}
    \end{subfigure}
    
    \begin{subfigure}{0.8\linewidth}
    \includegraphics[width=\linewidth]{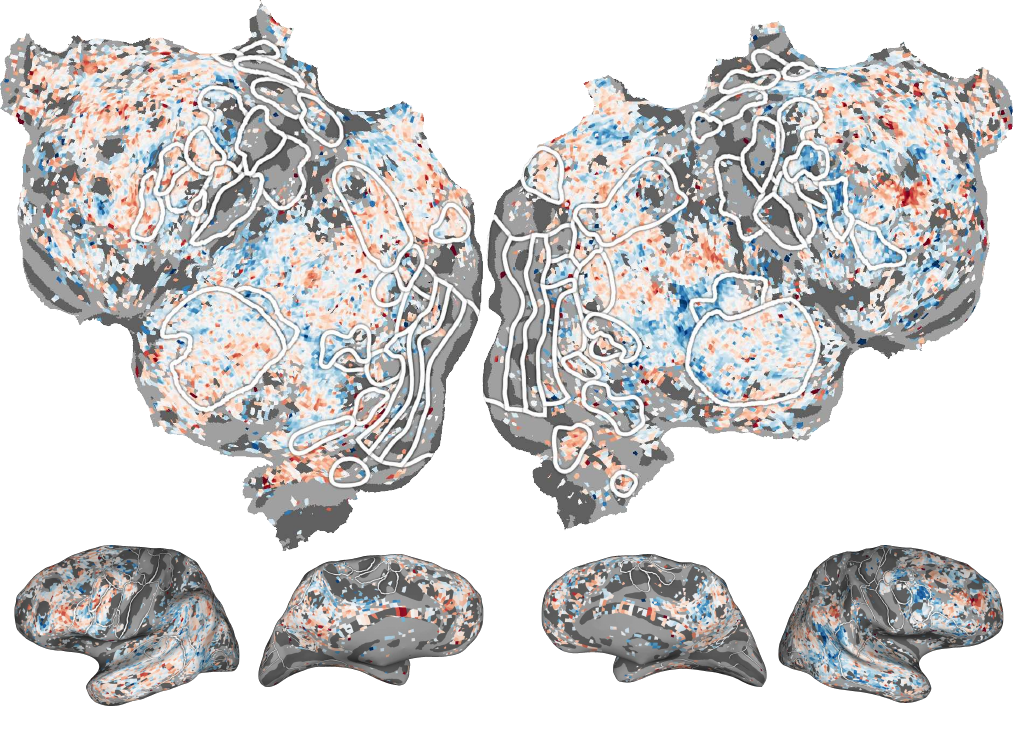} 
    \caption{S02}
    \end{subfigure}
    
    \caption{Difference in Encoding Performance (Right Hem. Fine-tuning -- Left Hem. Fine-tuning) for subjects S01 and S02. Subject S03 is shown in Fig.~\ref{fig:roi-encperf}C.}
    \label{fig:app-hemis-flatmaps}
\end{figure}

\newpage
\section{Per-subject probing results}
\label{app:subject-probing}

Here we show the un-averaged acoustic and semantic probe performance from Fig.~\ref{fig:probing}.

\begin{figure}[H]
    \centering
    \begin{subfigure}{0.55\linewidth}
    \includegraphics[width=\linewidth]{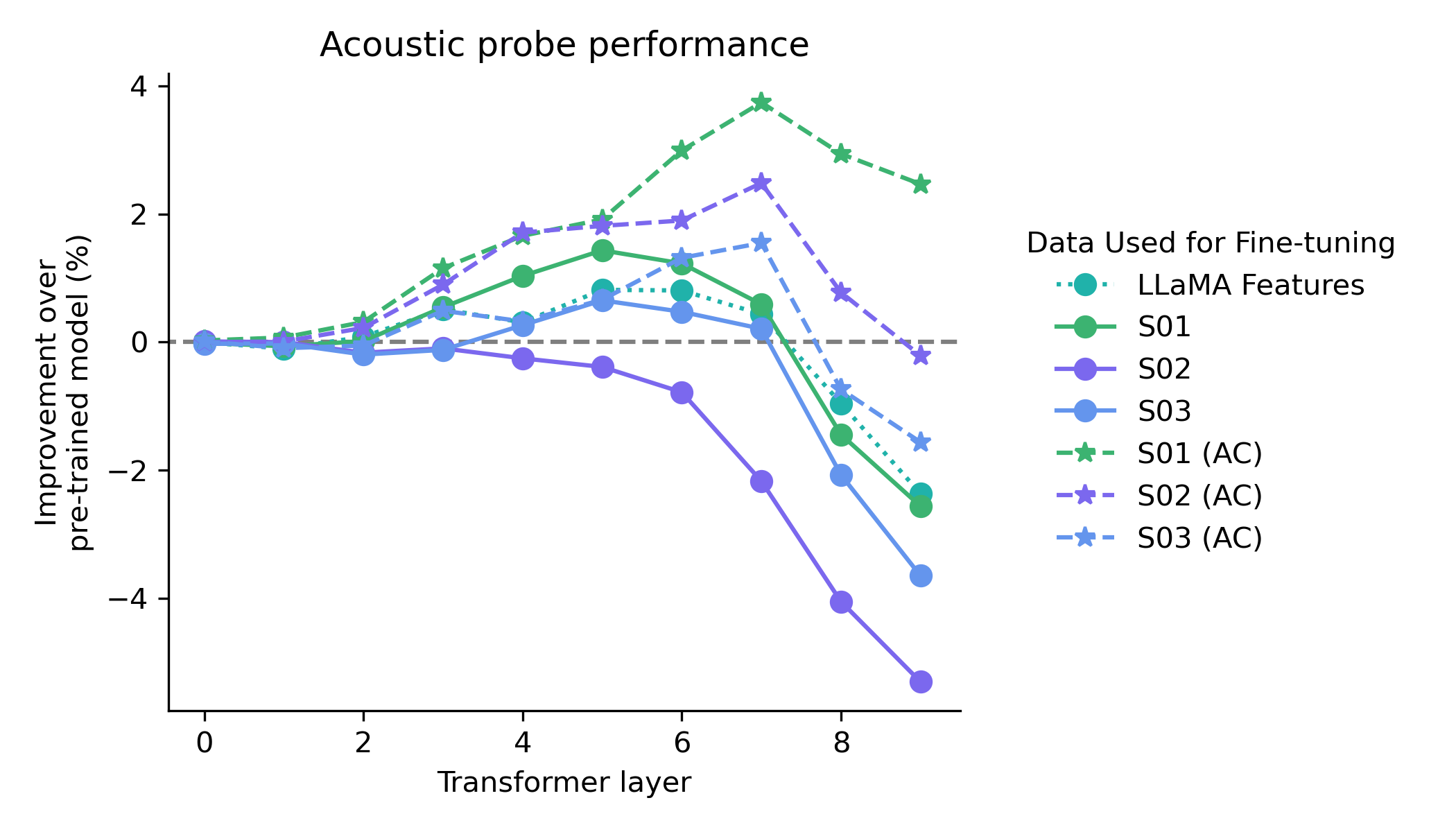} 
    \end{subfigure}
    
    \begin{subfigure}{0.55\linewidth}
    \includegraphics[width=\linewidth]{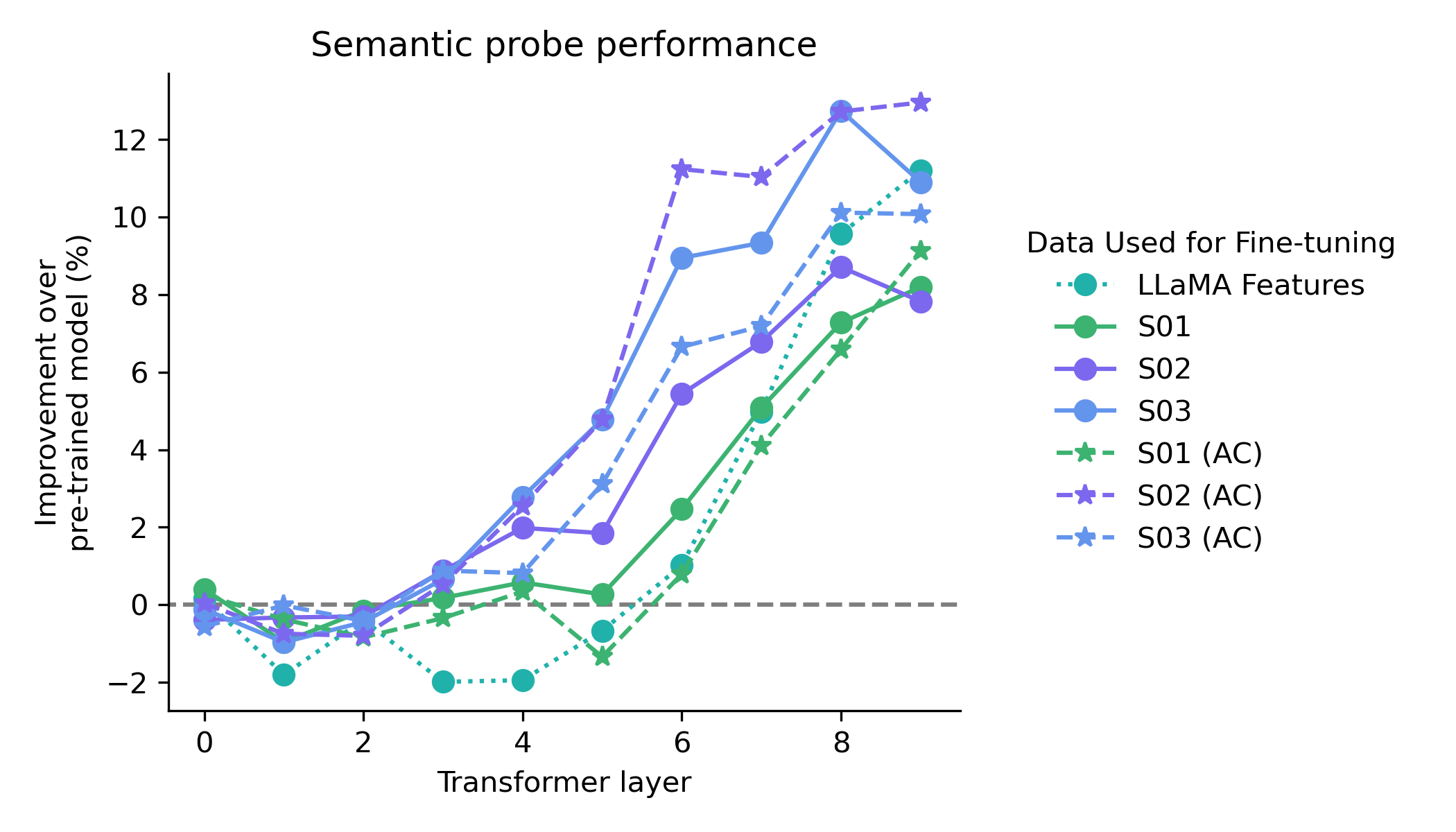} 
    \end{subfigure}
    
    \caption{Acoustic and semantic probe performance, shown separately for WavLM models fine-tuned on LLaMA, all of cortex, and only on AC. Performance averaged across fMRI participants is presented in Fig.~\ref{fig:probing}.}
    \label{fig:app-subject-probing}
\end{figure}

\end{document}